\documentclass[journal]{IEEEtran}
\usepackage{amsmath,amsfonts,amssymb}
\usepackage{graphicx}
\usepackage{cite}
\usepackage{booktabs}
\usepackage{algorithm}
\usepackage{algorithmic}
\usepackage{url}
\usepackage{multirow}
\usepackage{color}
\usepackage{subcaption}

\usepackage{hyperref}
\hypersetup{
    colorlinks=true,
    linkcolor=blue,
    citecolor=blue,
    urlcolor=blue,
}

\begin{document}

% =================================================================================
% TITLE AND ABSTRACT
% =================================================================================

% [MODIFIED] Title to fit the "Generative Simulation" theme
\title{GAC-KAN: An Ultra-Lightweight GNSS Interference Classifier for GenAI-Powered Consumer Edge Devices}

\author{Zhihan Zeng, \IEEEmembership{Graduate Student Member, IEEE}, Kaihe Wang, \IEEEmembership{Graduate Student Member, IEEE}, \\
Zhongpei Zhang, Yue Xiu, \IEEEmembership{Member, IEEE}

\thanks{Zhihan Zeng, Yue Xiu and Zhongpei Zhang are with the National Key Laboratory of Wireless Communications, University of Electronic Science and Technology of China (UESTC), Chengdu 611731, China (E-mail: 202511220608@std.uestc.edu.cn, xiuyue12345678@163.com, zhangzp@uestc.edu.cn). Kaihe Wang is with the University of Electronic Science and Technology of China (UESTC), Chengdu 611731, China (E-mail: khewang@yeah.net). The corresponding author is Zhongpei Zhang.}}
\maketitle

% [MODIFIED] Abstract to align with GenAI Special Track
\begin{abstract}
The integration of Generative AI (GenAI) into Consumer Electronics (CE)—from AI-powered assistants in wearables to generative planning in autonomous Uncrewed Aerial Vehicles (UAVs)—has revolutionized user experiences. However, these GenAI applications impose immense computational burdens on edge hardware, leaving strictly limited resources for fundamental security tasks like Global Navigation Satellite System (GNSS) signal protection. Furthermore, training robust classifiers for such devices is hindered by the scarcity of real-world interference data. 

To address the dual challenges of data scarcity and the extreme efficiency required by the GenAI era, this paper proposes a novel framework named GAC-KAN. First, we adopt a physics-guided simulation approach to synthesize a large-scale, high-fidelity jamming dataset, mitigating the data bottleneck. Second, to reconcile high accuracy with the stringent resource constraints of GenAI-native chips, we design a Multi-Scale Ghost-ACB-Coordinate (MS-GAC) backbone. This backbone combines Asymmetric Convolution Blocks (ACB) and Ghost modules to extract rich spectral-temporal features with minimal redundancy. 

Replacing the traditional Multi-Layer Perceptron (MLP) decision head, we introduce a Kolmogorov-Arnold Network (KAN), which employs learnable spline activation functions to achieve superior non-linear mapping capabilities with significantly fewer parameters. Experimental results demonstrate that GAC-KAN achieves an overall accuracy of 98.0\%, outperforming state-of-the-art baselines. Significantly, the model contains only 0.13 million parameters—approximately 660 times fewer than Vision Transformer (ViT) baselines. This extreme lightweight characteristic makes GAC-KAN an ideal "always-on" security companion, ensuring GNSS reliability without contending for the computational resources required by primary GenAI tasks.
\end{abstract}

% [MODIFIED] Keywords to include Generative Simulation

% =================================================================================
% SECTION I: INTRODUCTION
% =================================================================================
\section{Introduction}
\label{sec:intro}

% [MODIFIED] Introduction to focus on the conflict between GenAI and Security Resources
\IEEEPARstart{T}{he} current landscape of Consumer Electronics (CE) is being reshaped by the advent of Generative AI (GenAI). From interactive fashion design tools utilizing diffusion models \cite{Yan2025SketchFit} to smartphones running on-device Large Language Models (LLMs), the demand for intelligent edge processing is unprecedented. This shift towards pervasive AI on resource-constrained devices, often referred to as TinyML, promises ultra-low latency and enhanced privacy \cite{Jhaveri2024TinyML}. However, the reliability of these advanced AI systems fundamentally depends on the integrity of their sensory inputs and decision-making processes under uncertainty \cite{Feng2025Intent}, particularly regarding Positioning, Navigation, and Timing (PNT) services provided by Global Navigation Satellite Systems (GNSS). If a GenAI-driven drone loses its location due to interference, its generative capabilities become futile or even dangerous.

Despite the critical need for GNSS security, a significant conflict arises in hardware resource allocation. GenAI models are computationally intensive, consuming the vast majority of the Neural Processing Unit (NPU) and battery budget on consumer devices \cite{mehr2025towards}. This creates a dilemma: the device needs robust interference monitoring, but efficient power allocation is critical to maintain network longevity \cite{Yin2024GWO}. Traditional Deep Learning (DL) classifiers (like ResNet or ViT) are often too heavy to run concurrently with GenAI applications. Therefore, next-generation consumer electronics demand \textit{ultra-lightweight} security frameworks that can operate in the background without throttling the primary GenAI tasks. Recent works have demonstrated the viability of such lightweight approaches in other domains, such as fuzzy-driven intrusion detection for IoMT \cite{Aljuhani2025Lightweight} and Riemannian manifold-based detection for wearables \cite{Liu2026Lightweight}, proving that high accuracy is achievable with minimal computational overhead.

Another hurdle in deploying AI for interference classification is the scarcity of labeled data. Unlike GenAI domains where vast text or image datasets exist, real-world jamming signals are difficult to capture in diverse scenarios. This necessitates a generative simulation approach, where high-fidelity synthetic data is generated based on physical signal models to train robust classifiers \cite{wang2018gnss}.

To address these challenges—efficient co-existence with GenAI workloads and data scarcity—this paper proposes the GAC-KAN framework. This architecture integrates a Multi-Scale Ghost-ACB-Coordinate backbone with a Kolmogorov-Arnold Network (KAN) classification head. We employ Ghost modules \cite{han2020ghostnet} and Asymmetric Convolution Blocks (ACB) \cite{ding2019acnet} to reduce redundancy, while utilizing the learnable activation functions of KANs \cite{liu2024kan} for superior generalization with fewer parameters. This design aligns with the strict energy-efficiency requirements of next-generation consumer electronics and Green Communications \cite{nguyen20226g}.

The main contributions of this study are as follows:
\begin{itemize}
    \item We address the data scarcity issue by utilizing a physics-guided generative simulation strategy, creating a comprehensive dataset for jamming primitives.
    \item We introduce a lightweight feature extraction backbone tailored for edge devices, integrating asymmetric convolutions and ghost modules to effectively reduce computational complexity.
    \item We apply KANs to GNSS interference classification, demonstrating that learnable activation functions on network edges offer superior generalization compared to traditional perceptrons, specifically benefiting compact models.
    \item Extensive simulations show that the proposed framework achieves state-of-the-art accuracy (98.0\%) with minimal computational overhead (0.13 M parameters), providing a practical solution for always-on interference monitoring alongside GenAI applications.
\end{itemize}

The remainder of this paper is organized as follows. Section \ref{sec:related_work} reviews related work. Section \ref{sec:system_model} details the system model and generative signal formulations. Section \ref{sec:methodology} describes the proposed lightweight network architecture. Section \ref{sec:Simulation Results and Analysis} presents the simulation results and performance analysis. Finally, Section \ref{sec:conclusion} concludes the paper.

% =================================================================================
% SECTION II: RELATED WORK
% =================================================================================
\section{Related Work}
\label{sec:related_work}

The evolution of GNSS interference mitigation has transitioned from statistical signal processing to data-driven deep learning. With the growing demand for intelligent, autonomous, and resource-efficient interference recognition on consumer edge devices \cite{nguyen20226g}, the focus has shifted towards embedded AI. This section reviews the literature across conventional processing, deep learning paradigms, and emerging lightweight architectures suitable for consumer applications.

\subsection{Conventional Signal Processing Approaches}
Traditional interference defense relying on statistical distinctiveness remains common in legacy receivers. Time-frequency (TF) analysis and filtering are the cornerstones of these methods. For instance, statistical inference based on the Chi-square goodness-of-fit test and variance analysis in the TF domain has been widely adopted to detect anomalies \cite{wang2018gnss, wang2017time}. To mitigate specific narrowband or swept-frequency interference, Adaptive Notch Filters (ANF) are commonly employed in low-cost receivers due to their low implementation cost \cite{borio2014multi, gamba2019performance, qin2020assessment}. However, ANFs often introduce signal distortion, necessitating complex bias compensation which strains the processing budget of consumer chips.

For non-stationary signals like Chirp jamming, transform-domain methods are preferred. The FrFT has shown superior energy concentration properties for chirp signals \cite{sun2024gnss, sun2024novel, alvarez2025chirp}. Similarly, Luo \textit{et al.} \cite{luo2024zak} utilized the Zak transform to exploit the sparsity of interference in the Doppler-shift domain. Other approaches include Wavelet transforms for multi-scale analysis \cite{dovis2011use} and Nonnegative Matrix Factorization (NMF) \cite{dasilva2023radio}. Despite their theoretical robustness, these methods often require manual parameter tuning and struggle to distinguish complex compound interference patterns without prior knowledge, limiting their "plug-and-play" capability in consumer electronics.

\subsection{Deep Learning-Based Interference Recognition}
To overcome the limitations of model-based methods, DL has been extensively applied to treat interference recognition as a pattern classification problem. Early adoption of CNNs demonstrated significant performance gains over Support Vector Machines (SVMs) by automatically extracting features from spectrograms \cite{ferre2019jammer, liu2019deep}. Recent advancements have focused on handling complex scenarios. For example, Xiao \textit{et al.} \cite{xiao2025compound} introduced a dual-stream network (TFPENet), while Li \textit{et al.} \cite{li2025gnss} utilized Dual GCN. For object-detection-based approaches, YOLOv5 has been adapted to detect multiple interference sources simultaneously \cite{liu2024gnss}.

Although these heavy-weight models achieve high accuracy, they typically involve millions of parameters and high FLOPs. This poses a significant challenge for consumer GNSS receivers, which often share computational resources with other application processes on a System-on-Chip (SoC). Efforts to reduce complexity include using Gated Recurrent Units (GRU) \cite{mehr2025towards} and optimizing classical machine learning synergies \cite{vandermerwe2024optimal}, yet achieving a balance between high accuracy and the extreme latency/power constraints of edge devices remains an open challenge.

\subsection{Lightweight Architectures and Kolmogorov-Arnold Networks}
The necessity for edge-deployable AI has driven research into lightweight neural primitives. Prior works have explored dynamic kernel selection, such as Selective Kernel (SK) Networks \cite{li2019selective, li2023large}, to adaptively adjust receptive fields, which has proven effective in modulation recognition tasks \cite{yang2023selective}. However, these mechanisms often incur additional branching latency. To strictly control computational cost, Han \textit{et al.} \cite{han2020ghostnet} proposed GhostNet, which generates redundant feature maps via cheap linear operations. Similarly, Ding \textit{et al.} \cite{ding2019acnet} introduced Asymmetric Convolution Blocks (ACNet) to strengthen kernel skeletons during training while fusing them into standard kernels for inference, incurring no extra deployment cost—an ideal characteristic for firmware-constrained devices.

A paradigm shift in neural architecture is the emergence of KANs \cite{liu2024kan}. Unlike MLPs that use fixed activation functions on nodes, KANs employ learnable spline functions on edges, offering superior parameter efficiency. Very recently, Jia \textit{et al.} \cite{jia2025lightweight} explored a lightweight classifier combining Multi-Scale Feature Fusion with KAN (MSFF-KAN). Building upon these innovations, our work proposes a novel framework that integrates the structural efficiency of Ghost-ACB modules with the non-linear representational power of KANs, aiming to establish a new benchmark for energy-efficient interference recognition in consumer electronics.

% =================================================================================
% SECTION III: SYSTEM MODEL
% =================================================================================
\section{System Model}
\label{sec:system_model}

This section details the signal formulation for the GNSS receiver in the presence of interference, defines the mathematical models for the specific jamming primitives considered in this study, and describes the time-frequency transformation used for feature extraction.

    \subsection{Signal Formulation}
    We consider a typical GNSS receiver architecture operating in an environment characterized by additive noise and intentional interference. The received RF signal $y(t)$ at the antenna input can be modeled as:
    \begin{equation}
        y(t) = s_{\text{GNSS}}(t) + J(t) + n(t),
    \end{equation}
    where $s_{\text{GNSS}}(t)$ represents the authentic satellite navigation signal, $J(t)$ denotes the jamming interference signal, and $n(t)$ is the additive white Gaussian noise (AWGN) with zero mean and variance $\sigma_n^2$.

    Given that GNSS signals are received at extremely low power levels (typically below the thermal noise floor) \cite{zhang2011effect, wesson2018gnss}, and interference recognition systems are primarily concerned with high-power jamming scenarios, the received signal is dominated by the interference and noise components. Consequently, for the purpose of interference classification, the signal model is effectively dominated by the JNR. The discretized received signal $y[n]$ sampled at frequency $f_s$ is given by
    \begin{equation}
        y[n] = y(n T_s) = J[n] + \eta[n], \quad n = 0, 1, \dots, N-1,
    \end{equation}
    where $T_s = 1/f_s$ is the sampling interval, $N$ is the number of samples per observation window, and $\eta[n]$ represents the complex discrete noise component \cite{chen2022gnss, mehr2025deep}.

    % =================================================================================
    % FIGURE: JAMMING SPECTROGRAMS
    % =================================================================================
    \begin{figure}[!t]
        \centering
        % --- Row 1 ---
        \includegraphics[width=0.32\linewidth]{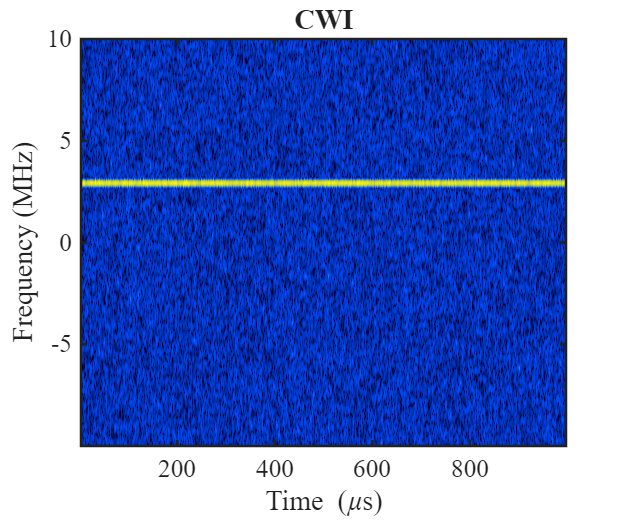}
        \hfil
        \includegraphics[width=0.32\linewidth]{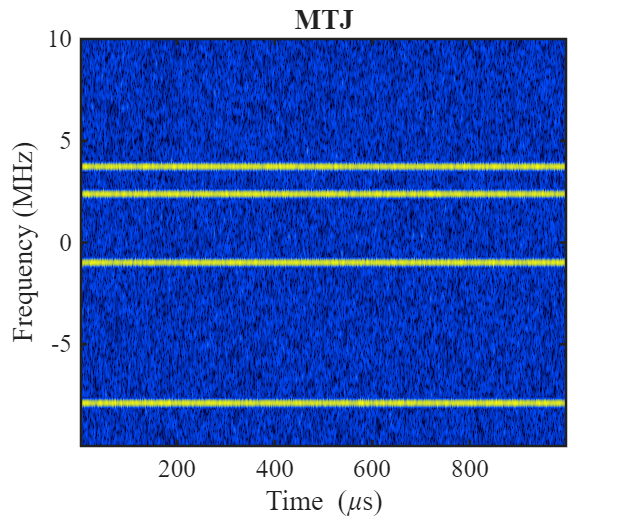}
        \hfil
        \includegraphics[width=0.32\linewidth]{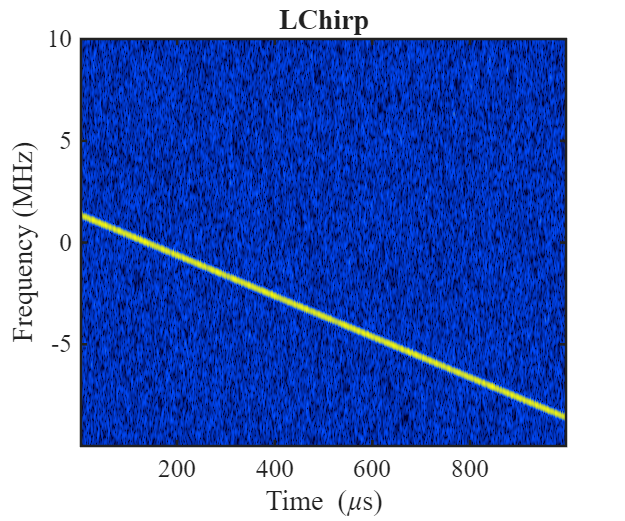}
        
        \vspace{0pt} 
        
        % --- Row 2 ---
        \includegraphics[width=0.32\linewidth]{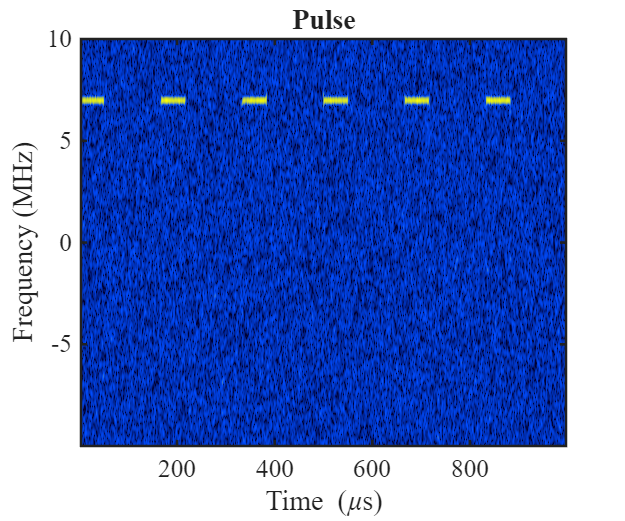}
        \hfil
        \includegraphics[width=0.32\linewidth]{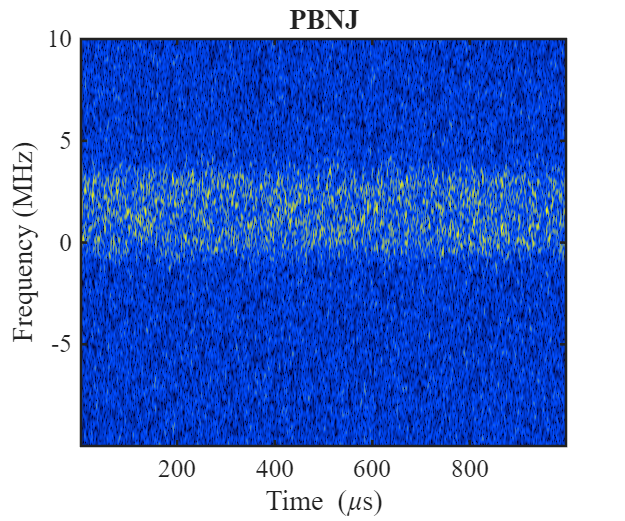}
        \hfil
        \includegraphics[width=0.32\linewidth]{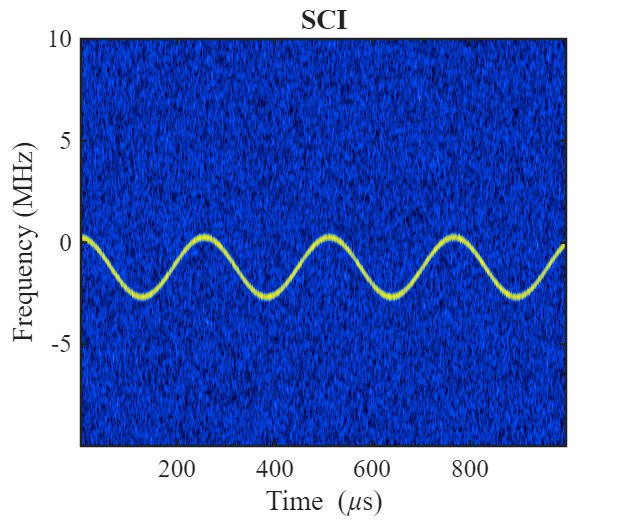}
        
        \vspace{0pt} 
        
        % --- Row 3 (Centered) ---
        \includegraphics[width=0.32\linewidth]{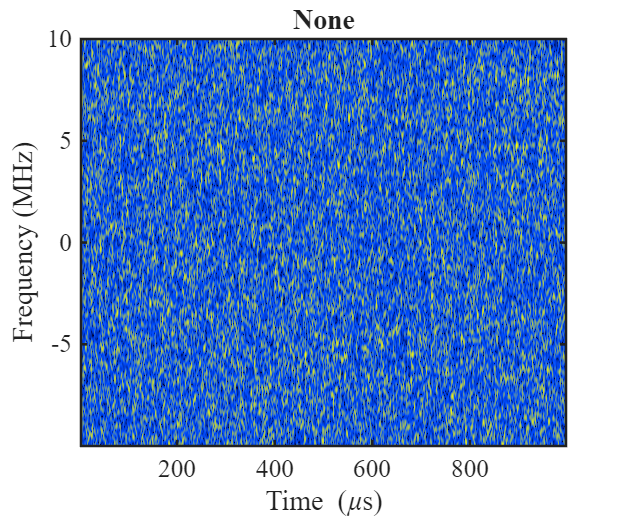}
        
        \caption{Time-Frequency representations (spectrograms) of the generated jamming primitives considered in the system model. The horizontal axis represents time ($\mu s$) and the vertical axis represents frequency (MHz).}
        \label{fig:jamming_types}
    \end{figure}

\subsection{Fundamental Jamming Primitives}
    To evaluate the robustness of the proposed framework, we consider six fundamental signal classes \cite{ferre2019jammer, wang2018gnss}. These classes cover distinct time-frequency characteristics, ranging from stationary narrow-band signals to non-stationary wide-band modulations, as illustrated in Fig. \ref{fig:jamming_types}.

        \subsubsection{Single-Tone Jamming (STJ)}
        Single-Tone Jamming, also referred to as Continuous Wave Interference (CWI), concentrates high power on a single frequency, effectively disrupting the receiver's tracking loops \cite{borio2014multi}. It is modeled as:
        \begin{equation}
            J_{\text{STJ}}(t) = \sqrt{P_J} e^{j(2\pi f_c t + \phi)},
        \end{equation}
        where $P_J$ is the jamming power, $f_c$ is the carrier frequency randomly located within the receiver bandwidth, and $\phi$ is a random initial phase uniformly distributed in $[0, 2\pi)$.

        \subsubsection{Multi-Tone Jamming (MTJ)}
        Multi-Tone Jamming consists of a superposition of $K$ distinct single-tone signals distributed across the frequency band \cite{wang2017time}. This technique targets multiple sub-bands simultaneously and is defined as:
        \begin{equation}
            J_{\text{MTJ}}(t) = \sum_{k=1}^{K} \sqrt{P_k} e^{j(2\pi f_k t + \phi_k)},
        \end{equation}
        where $P_k$, $f_k$, and $\phi_k$ correspond to the power, frequency, and phase of the $k$-th tone, respectively. The frequencies $\{f_k\}$ are typically independent and randomly distributed.

        \subsubsection{Linear Frequency Modulation (LFM)}
        LFM jamming, often called Chirp interference \cite{sun2024gnss, alvarez2025chirp}, sweeps the jamming frequency linearly over a specific bandwidth $B_{\text{sweep}}$ within a period $T_{\text{sweep}}$. This non-stationary signal creates a sliding effect in the time-frequency domain defined as follows:
        \begin{equation}
            J_{\text{LFM}}(t) = \sqrt{P_J} e^{j 2\pi (f_0 t + \frac{1}{2} k t^2)},
        \end{equation}
        where $f_0$ is the starting frequency and $k = B_{\text{sweep}}/T_{\text{sweep}}$ represents the chirp rate (frequency slope). The sign of $k$ determines whether the chirp is an up-chirp or down-chirp.

        \subsubsection{Pulse Jamming}
        Pulse jamming is characterized by intermittent transmission, defined by a series of RF pulses \cite{garzia2021sub}. This allows the jammer to achieve high peak power while maintaining a lower average power. The signal is modeled as:
        \begin{equation}
            J_{\text{Pulse}}(t) = p(t) \cdot \sqrt{P_J} e^{j(2\pi f_c t + \phi)},
        \end{equation}
        where $p(t)$ is a rectangular pulse train function defined by the Pulse Repetition Interval (PRI) and Pulse Width (PW). The duty cycle is determined by the ratio $\text{PW}/\text{PRI}$.
        
\begin{figure*}[!t]
    \centering
    \includegraphics[width=0.95\linewidth]{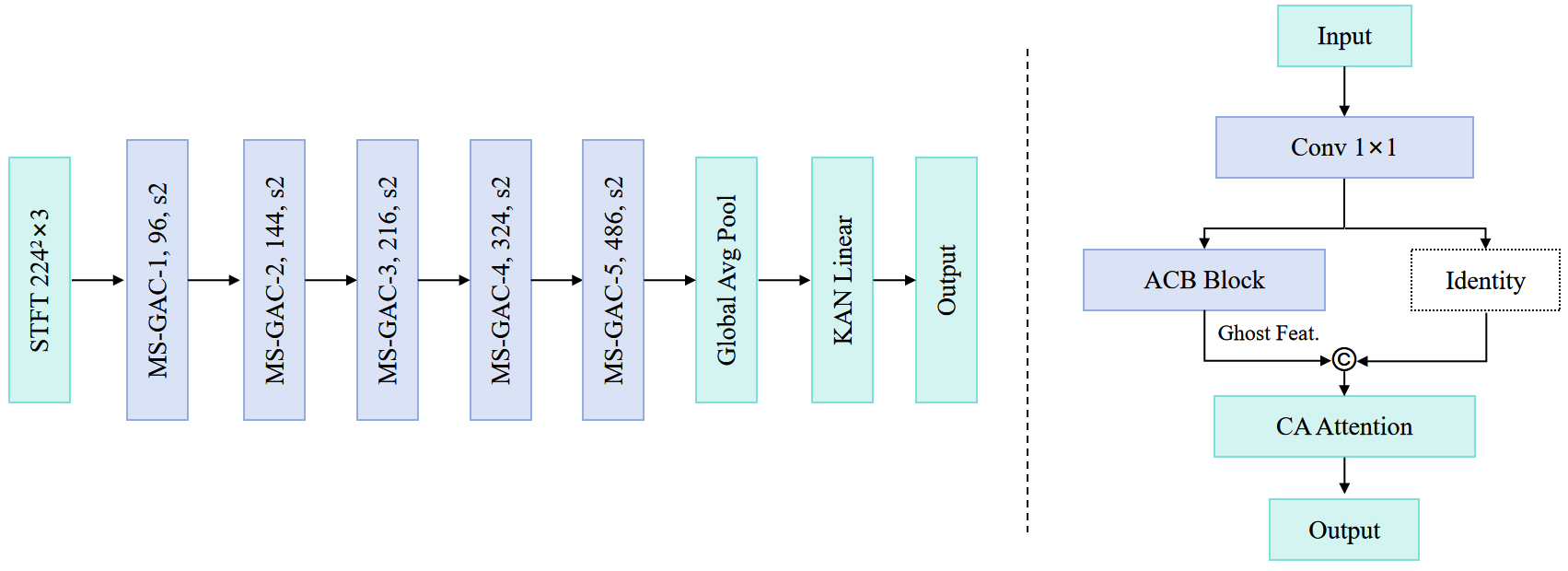}
    \caption{The overall architecture of the proposed GAC-KAN (left) and the detailed structure of the Ghost-ACB-CA Unit (right). The network processes STFT spectrograms through stacked MS-GAC blocks and employs a KAN head for classification.}
    \label{fig:network_overview}
\end{figure*}

        \subsubsection{Partial-Band Noise Jamming (PBNJ)}
        Unlike single-tone or chirp jamming which are deterministic, PBNJ is a stochastic process \cite{dasilva2023radio}. It is generated by band-limiting white Gaussian noise to a specific bandwidth $B_J$ significantly smaller than the full receiver bandwidth $B_R$, and modulating it to a carrier frequency $f_c$ defined as follows:
        \begin{equation}
            J_{\text{PBNJ}}(t) = [w(t) * h_{\text{LPF}}(t)] e^{j 2\pi f_c t},
        \end{equation}
        where $w(t)$ is complex white Gaussian noise, $h_{\text{LPF}}(t)$ is the impulse response of a low-pass filter with cutoff frequency corresponding to $B_J/2$, and $*$ denotes the convolution operation. In the time-frequency domain, this appears as a block of energy spanning the duration of the signal but restricted in frequency.
        
        \subsubsection{Sinusoidal Chirp Interference (SCI)}
        SCI introduces a non-linear frequency variation \cite{xiao2025compound}. The instantaneous frequency varies sinusoidally over time, which can defeat jamming mitigation algorithms designed for linear sweeps. It is formulated as:
        \begin{equation}
            J_{\text{SCI}}(t) = \sqrt{P_J} e^{j(2\pi f_c t + \beta \sin(2\pi f_{\text{mod}} t))},
        \end{equation}
        where $f_{\text{mod}}$ is the modulation frequency and $\beta$ is the modulation index.

\subsection{Time-Frequency Representation via STFT}
    Since the interference signals exhibit distinct characteristics in both time and frequency domains (e.g., LFM varies in frequency over time, while Pulse varies in amplitude over time), we utilize the Short-Time Fourier Transform (STFT) to generate a 2D time-frequency representation \cite{wang2018gnss, xiao2025compound}.
    
    The discrete STFT of the received signal $y[n]$ is computed using a sliding window function $w[n]$ defined as follows:
    \begin{equation}
        \text{STFT}\{y\}[m, k] = \sum_{n=0}^{N-1} y[n] w[n - mR] e^{-j \frac{2\pi}{N_{\text{fft}}} k n},
    \end{equation}
    where $m$ is the time index, $k$ is the frequency index, $R$ is the hop size (step length) between successive windows, and $N_{\text{fft}}$ is the number of FFT points.
    
    To obtain the input features for the deep learning model, we compute the logarithmic magnitude spectrogram $S[m, k]$:
    \begin{equation}
        S[m, k] = 20 \log_{10} \left( | \text{STFT}\{y\}[m, k] | + \epsilon \right),
    \end{equation}
    where $\epsilon$ is a small constant to prevent numerical instability. This transformation maps the 1D time-domain signal into a 2D image-like tensor, enabling the subsequent CNNs to extract spatial patterns corresponding to different jamming signatures \cite{cai2019jamming, liu2019deep, mehr2025towards}.

% =================================================================================
% SECTION V: METHODOLOGY
% =================================================================================
\section{Methodology}
\label{sec:methodology}

This section delineates the proposed lightweight deep learning framework designed for efficient GNSS interference recognition. We first formulate the classification problem and then detail the architectural components, including the asymmetric feature extraction backbone, the coordinate attention mechanism for time-frequency localization, and the spline-based Kolmogorov-Arnold classification head.

    \subsection{Problem Formulation}
    The objective of GNSS interference classification is to map the observed signal to a specific jamming category. Let $\mathcal{D} = \{(\mathbf{X}_i, y_i)\}_{i=1}^{N}$ denote the dataset consisting of $N$ labeled samples \cite{ferre2019jammer}. Here, $\mathbf{X}_i \in \mathbb{R}^{C \times H \times W}$ represents the preprocessed time-frequency spectrogram of the received signal, where $C$, $H$, and $W$ denote the channel depth, frequency bins, and time steps, respectively. The scalar $y_i \in \{1, \dots, K\}$ is the ground truth label corresponding to one of the $K$ jamming classes defined in Section \ref{sec:system_model}.

    We seek to learn a non-linear mapping function $\mathcal{F}: \mathbb{R}^{C \times H \times W} \to \mathbb{R}^{K}$, parameterized by weights $\Theta$, such that the predicted probability distribution $\mathbf{\hat{y}}_i = \mathcal{F}(\mathbf{X}_i; \Theta)$ minimizes the divergence from the true label $y_i$. The optimization objective is defined by minimizing the cross-entropy loss augmented with regularization terms to ensure generalization as follows:
    \begin{equation}
        \mathcal{L}(\Theta) = -\frac{1}{N} \sum_{i=1}^{N} \sum_{k=1}^{K} \mathbb{I}(y_i = k) \log(\hat{y}_{i,k}) + \lambda \mathcal{R}(\Theta),
    \end{equation}
    where $\mathbb{I}(\cdot)$ is the indicator function, $\hat{y}_{i,k}$ is the predicted probability for class $k$, and $\mathcal{R}(\Theta)$ represents the sparsity-inducing regularization applied to the network parameters.

\begin{figure*}[!t]
    \centering
    \includegraphics[width=\linewidth]{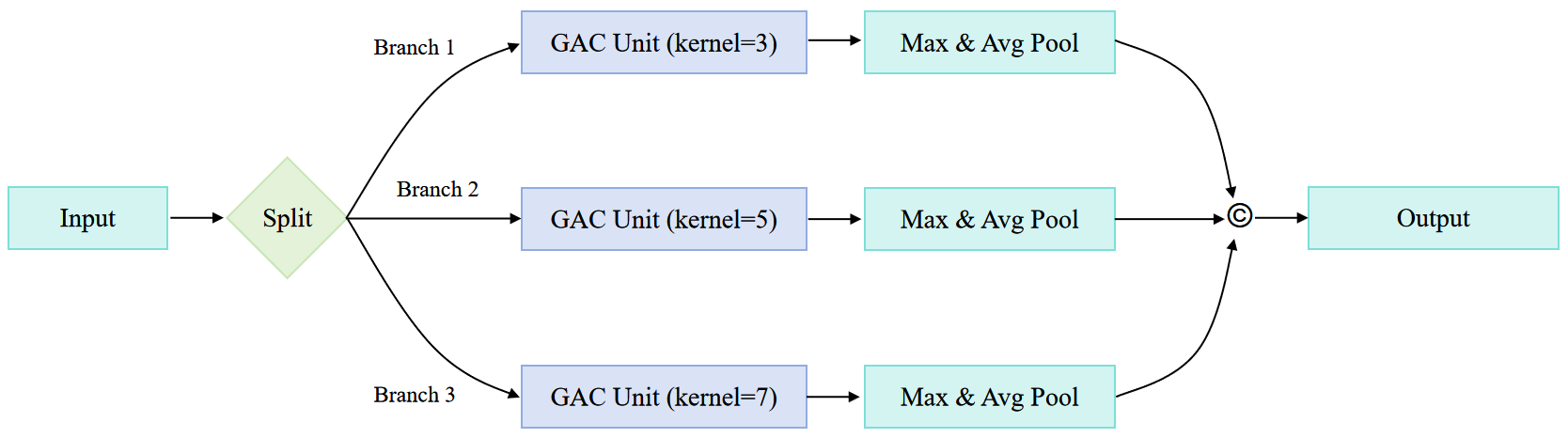}
    \caption{Structure of the Multi-Scale Ghost-ACB-Coordinate (MS-GAC) Block utilizing parallel branches with varying receptive fields.}
    \label{fig:ms_gac_block}
\end{figure*}

\subsection{Overall Architecture: GAC-KAN}
To address the challenges of deploying robust interference classification models on resource-constrained edge devices, we propose a lightweight network named GAC-KAN (Ghost-ACB-Coordinate-KAN). 

As illustrated in Fig.~\ref{fig:network_overview}, the network takes the STFT spectrogram as input and processes it through a stack of MS-GAC blocks. These blocks effectively capture multi-resolution spectral-temporal features while maintaining a low parameter count. Finally, a Global Average Pooling (GAP) layer compresses the feature maps, which are then classified by a KAN head, replacing the traditional MLP \cite{jia2025lightweight}.

\subsection{Multi-Scale Ghost-ACB-Coordinate (MS-GAC) Block}
Signal interference exhibits diverse characteristics in the time-frequency domain, ranging from narrow-band signals to wide-band frequency modulation. To capture these features simultaneously, we design the MS-GAC Block.

As shown in Fig.~\ref{fig:ms_gac_block}, the block employs a multi-branch Inception-style architecture. The input feature map $\mathbf{X}$ is split into three parallel branches, each processing the signal with a different kernel size $k \in \{3, 5, 7\}$. This allows the network to learn representations at different scales defined as follows:
\begin{equation}
    \mathbf{Y}_{branch}^{(k)} = \mathcal{F}_{GAC}(\mathbf{X}, k),
\end{equation}
where $\mathcal{F}_{GAC}$ denotes the operation of the Ghost-ACB-CA unit. To preserve both dominant features and background context, the outputs of all branches are processed by parallel Max Pooling and Average Pooling layers, which are subsequently concatenated to form the final output of the block.

\subsection{Ghost-ACB-CA Unit}
The core component of the MS-GAC block is the Ghost-ACB-CA unit, detailed in the right panel of Fig.~\ref{fig:network_overview}. It integrates three key technologies to balance efficiency and performance:

\subsubsection{Asymmetric Ghost Module}
Standard convolutional layers introduce redundancy. We adopt the Ghost module principle but enhance the cheap operation using ACB. The primary convolution generates intrinsic features \cite{han2020ghostnet}. To generate "ghost" features with rotational robustness and stronger central skeletons, we employ ACB, which sums three parallel kernels ($d\times d$, $d\times 1$, and $1\times d$) during training and fuses them into a single kernel during inference defined as follows \cite{ding2019acnet}:
\begin{equation}
    \mathbf{K}_{\text{fused}} = \mathbf{K}_{\text{square}} \oplus \mathbf{K}_{\text{ver}} \oplus \mathbf{K}_{\text{hor}}.
\end{equation}
This fusion ensures no additional computational cost during deployment compared to standard convolutions.

\subsubsection{Time-Frequency Coordinate Attention (CA)}
Jamming signals often localize in specific frequency bands or time slots. To explicitly model these inter-dependencies, we embed a Coordinate Attention (CA) mechanism immediately after the Ghost feature concatenation. Unlike SE-Net which uses global pooling, CA factorizes channel attention into two 1D feature encoding processes along the time ($W$) and frequency ($H$) directions.
The output feature map is re-weighted as:
\begin{equation}
    y_c(i, j) = x_c(i, j) \times g_c^h(i) \times g_c^w(j),
\end{equation}
where $g_c^h$ and $g_c^w$ are the attention vectors generated by aggregating features along the vertical and horizontal directions, respectively. This forces the network to focus on the exact coordinates of the interference energy.

\subsection{Kolmogorov-Arnold Classification Head}
Conventional CNNs typically use a linear MLP as the classifier. In GAC-KAN, we employ a KAN layer. Based on the Kolmogorov-Arnold representation theorem, KANs replace fixed linear weights with learnable non-linear activation functions on the edges \cite{liu2024kan}.
For an input vector $\mathbf{x}$, the activation function $\phi(x)$ on each edge is defined as a linear combination of a basis function $b(x)$ and a learnable B-spline $s(x)$ as follows:
\begin{equation}
    \phi(x) = w_b \text{SiLU}(x) + w_s \sum_{i} c_i B_i(x),
\end{equation}
where $w_b, w_s$ are scaling factors and $c_i$ are learnable coefficients.

\subsubsection{Regularized Loss Function}
To further enforce sparsity and prevent overfitting in the KAN layer, we incorporate an $L_1$ regularization penalty on the spline coefficients. The total training objective is defined as follows:
\begin{equation}
    \mathcal{L}_{total} = \mathcal{L}_{CE}(\mathbf{y}, \hat{\mathbf{y}}) + \lambda \sum_{l=1}^{L} ||\mathbf{c}_l||_1,
\end{equation}
where $\mathcal{L}_{CE}$ is the Cross-Entropy loss (with label smoothing and Mixup augmentation as implemented), $\mathbf{c}_l$ represents the spline coefficients of the KAN layer, and $\lambda$ controls the regularization strength (set to $10^{-5}$ in our experiments).

% =================================================================================
% SECTION VI: SIMULATION RESULTS AND ANALYSIS
% =================================================================================
\section{Simulation Results and Analysis}
\label{sec:Simulation Results and Analysis}

\subsection{Dataset Generation and Preprocessing}
    To ensure a comprehensive evaluation of the proposed framework, a large-scale synthetic dataset was generated simulating a realistic GNSS reception environment. The simulation parameters were configured with a sampling rate $f_s$ of 20 MHz and a signal duration $T$ of 1 ms, resulting in a sequence length $N$ of 20,000 samples per observation. The dataset encompasses seven distinct categories: single-tone jamming, multi-tone jamming, linear frequency modulation, pulse jamming, partial-band noise jamming, sinusoidal chirp interference, and a noise-only class representing the interference-free scenario.

    The robustness of the model was evaluated across a wide dynamic range of JNRs. The JNR was varied from -25 dB to 10 dB in 5 dB increments. For each class and each JNR level, 1,000 Monte Carlo trials were conducted, yielding a total dataset size of 56,000 samples. The dataset was subsequently partitioned into training, validation, and testing subsets with a ratio of 70\%, 15\%, and 15\% respectively.

    Feature extraction was performed using the STFT. A Hann window of length 128 was applied with an overlap of 92\% between adjacent frames, and the FFT size was set to 4096 points to ensure high frequency resolution. The resulting complex spectrogram was converted to a logarithmic power scale. To enhance the visual distinctiveness of low-power features, a gamma correction factor of 0.9 was applied to the magnitude response. Finally, the spectrograms were pseudo-colored using a custom blue-to-yellow colormap and resized to a resolution of $224 \times 224$ pixels with three standard RGB channels, serving as the input tensor for the deep learning model.
\subsection{Implementation Details}
    
        \subsubsection{Hardware and Software Environment}
        Hardware specifications for this study included an Intel Core i9-12900KF CPU, 64 GB of system memory, and an NVIDIA GeForce RTX 5060 Ti (16 GB) GPU. The algorithms were coded in Python 3.11 using the PyTorch 2.x deep learning library, leveraging CUDA 11.8 and cuDNN for computational efficiency.

        \subsubsection{Training Strategy}
        The network parameters were optimized using the AdamW algorithm with an initial learning rate of $1 \times 10^{-3}$ and a weight decay of $1 \times 10^{-4}$ to mitigate overfitting. The training process spanned 200 epochs. A hybrid learning rate schedule was employed, consisting of a linear warmup phase for the first 5 epochs followed by a cosine annealing strategy for the remaining duration.

        To further enhance the generalization capability of the model, several regularization techniques were integrated. The Mixup data augmentation method was applied with a beta distribution parameter $\alpha = 1.0$, which generates synthetic training examples by linearly interpolating between pairs of inputs and their corresponding labels. Additionally, label smoothing with a factor of 0.1 was utilized to prevent the network from becoming over-confident in its predictions.

        The loss function was formulated as a combination of the soft-target cross-entropy loss derived from the Mixup strategy and an $L_1$ regularization term applied specifically to the spline coefficients of the KAN layer. This sparsity-inducing penalty, weighted by a factor of $\lambda = 1 \times 10^{-5}$, encourages the model to select the most informative basis functions, thereby reducing model complexity. The batch size was set to 64, and the model with the highest validation accuracy was retained for testing.

        % [MODIFIED] Subsection Title and Content to emphasize Generative Nature
        \subsubsection{Generative Signal Simulation and Parameterization}
        To overcome the lack of open-source interference datasets, we employ a physics-based generative simulation strategy to construct a large-scale synthetic dataset. This approach allows us to cover a diverse range of jamming primitives rather than relying on limited real-world captures.
        
        The jamming signals were synthesized based on rigorous mathematical models with randomized parameters to simulate diverse operational conditions. For STJ, the carrier frequency was uniformly distributed within 95\% of the receiver bandwidth. MTJ was constructed by superimposing 3 to 5 distinct tones with random phases and frequencies.

        The LFM signal was characterized by a sweep bandwidth of 10 MHz, covering half of the sampling rate, with the starting frequency randomized within the available band. Pulse jamming was modeled using a train of 6 pulses distributed over the signal duration, where the pulse width was set to 30\% of the pulse repetition interval.

        PBNJ was generated by filtering white Gaussian noise using a 4th-order Butterworth low-pass filter. The bandwidth of the noise was randomized between 10\% and 25\% of the sampling rate, and the center frequency was adjusted to ensure the jamming energy remained within the passband. SCI incorporated a non-linear frequency modulation, where the modulation frequency ranged from 10 kHz to 100 kHz and the modulation index $\beta$ was uniformly distributed between 10 and 50. This stochastic parameterization ensures that the trained model learns generalized features rather than memorizing specific signal instances.

\subsection{Performance Analysis}
\label{subsec:performance_analysis}

In this subsection, we present a comprehensive evaluation of the proposed GAC-KAN framework, comparing it against state-of-the-art baselines including the Vision Transformer (ViT-B/16), a standard Frequency-Attention CNN (FA-CNN), a Recurrent Convolutional Neural Network (RCNN), and a Spatio-Temporal Multi-Scale Fourier KAN (ST-MSFF-KAN).

    \subsubsection{Overall Comparison and Computational Efficiency}
    Table \ref{tab:performance_comparison} summarizes the overall classification accuracy (OA), model size (parameters), computational cost (FLOPs), and inference latency for all considered models on the test set.
    
    The proposed GAC-KAN achieves the highest overall accuracy of 98.0\%, outperforming the second-best method, ST-MSFF-KAN, by 1.2\%. Notably, our model demonstrates exceptional parameter efficiency. With only 0.13 M parameters, GAC-KAN is approximately 660 times smaller than the ViT-B/16 baseline (85.80 M parameters) and significantly lighter than the FA-CNN (57.05 M parameters). This drastic reduction in storage requirements is attributed to the replacement of dense MLP layers with the sparse, spline-based KAN layers and the use of the Ghost-ACB mechanism.
    
    In terms of computational complexity, GAC-KAN requires only 0.19 G FLOPs, the lowest among all compared methods. 
    % [Expert Note: Pivoted defense for inference time - Memory vs Speed tradeoff for Consumer Electronics]
    It is worth noting that the inference time of GAC-KAN (7.01 ms) is higher than that of simple CNN-based models like RCNN. However, in the context of consumer electronics, this presents a favorable trade-off. For human-interaction applications or autonomous navigation loops (typically operating at 10-50 Hz), a 7ms latency is well within the real-time operational margin. Conversely, the 660x reduction in model size is a critical advantage for firmware updates over the air (OTA) and storage on cost-constrained flash memory, which is often the primary bottleneck in mass-market devices.

    \begin{table}[htbp]
        \caption{Performance Comparison of Different Models}
        \label{tab:performance_comparison}
        \centering
        \begin{tabular}{lcccc}
            \toprule
            \textbf{Model} & \textbf{OA (\%)} & \textbf{Params (M)} & \textbf{FLOPs (G)} & \textbf{Time (ms)} \\
            \midrule
            \textbf{GAC-KAN} & \textbf{98.0} & \textbf{0.1289} & \textbf{0.1901} & 7.0140 \\
            ST-MSFF-KAN & 96.8 & 0.2883 & 0.5026 & 3.0140 \\
            ViT-B/16 & 95.2 & 85.8040 & 11.2855 & 6.6676 \\
            FA-CNN & 93.8 & 57.0530 & 0.7102 & 0.8693 \\
            RCNN & 91.8 & 13.1194 & 0.5547 & 0.3097 \\
            \bottomrule
        \end{tabular}
    \end{table}

    \subsubsection{Robustness Across JNR Regimes}
\begin{figure}[htbp]
        \centering
        \includegraphics[width=0.9\linewidth]{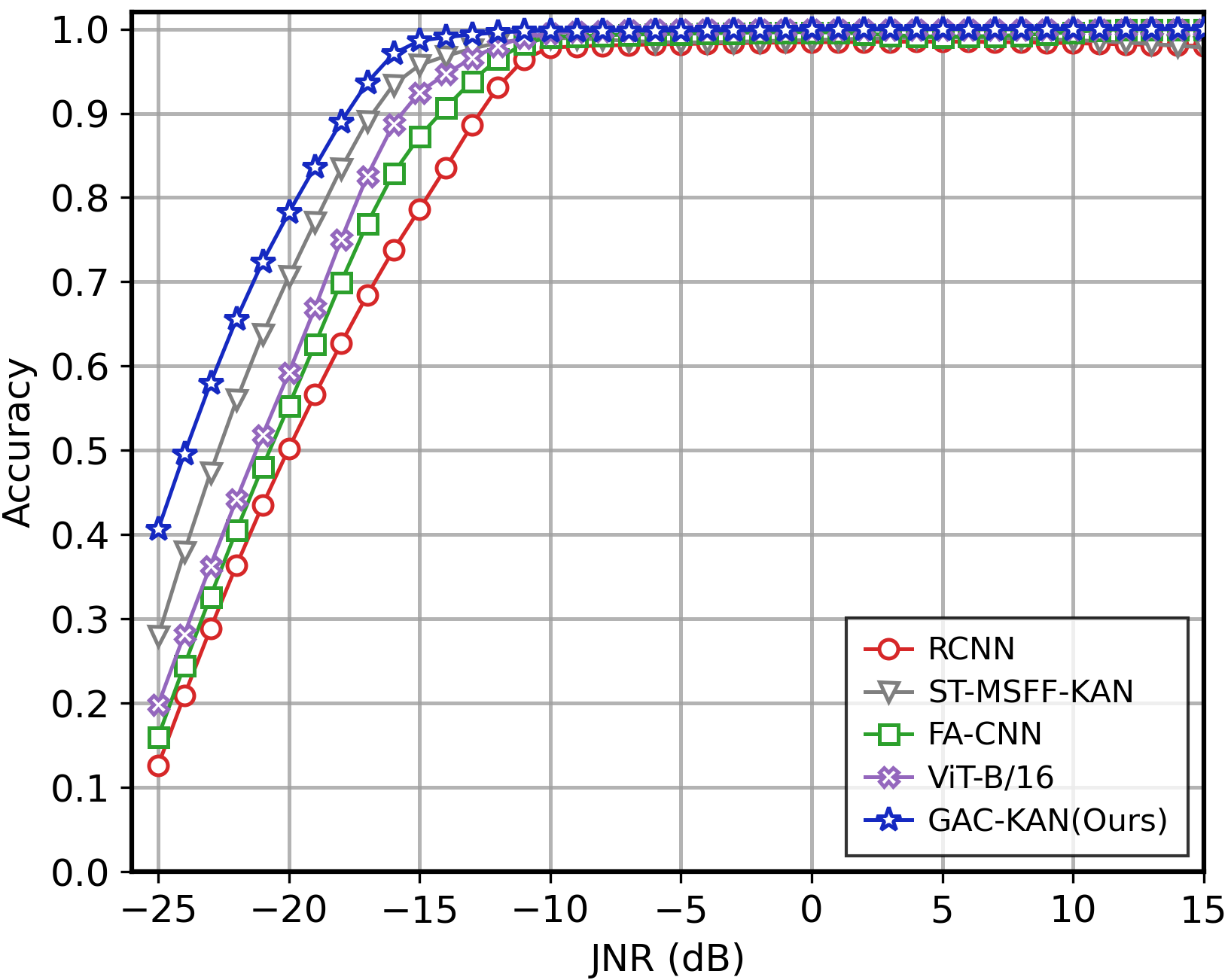}
        \caption{Classification accuracy versus Jamming-to-Noise Ratio (JNR). The proposed GAC-KAN demonstrates superior robustness in low JNR regimes (-25 dB to -15 dB).}
        \label{fig:accuracy_jnr}
    \end{figure}

    The reliability of interference recognition under low JNR conditions is critical for early warning systems. Fig. \ref{fig:accuracy_jnr} illustrates the classification accuracy of all models as a function of JNR, ranging from -25 dB to 15 dB.
    
    It is evident that all models achieve near-perfect accuracy ($>99\%$) when the JNR exceeds -10 dB. However, performance degrades rapidly for the baseline models as the JNR drops below -15 dB. The proposed GAC-KAN (depicted by the blue star curve) exhibits superior robustness in the low-SNR regime. Specifically, at a JNR of -20 dB, GAC-KAN maintains an accuracy of approximately 80\%, whereas the FA-CNN and RCNN drop to roughly 55\% and 50\%, respectively.
    
    This robustness can be attributed to the CA mechanism within our MS-GAC blocks, which effectively isolates the interference energy distribution in the time-frequency domain even when it is submerged in noise. Furthermore, the learnable activation functions in the KAN head provide a more flexible decision boundary compared to fixed ReLU activations used in standard CNNs.

\subsubsection{Inter-Class Discriminability Analysis} 
\begin{figure}[htbp]
        \centering
        \includegraphics[width=0.9\linewidth]{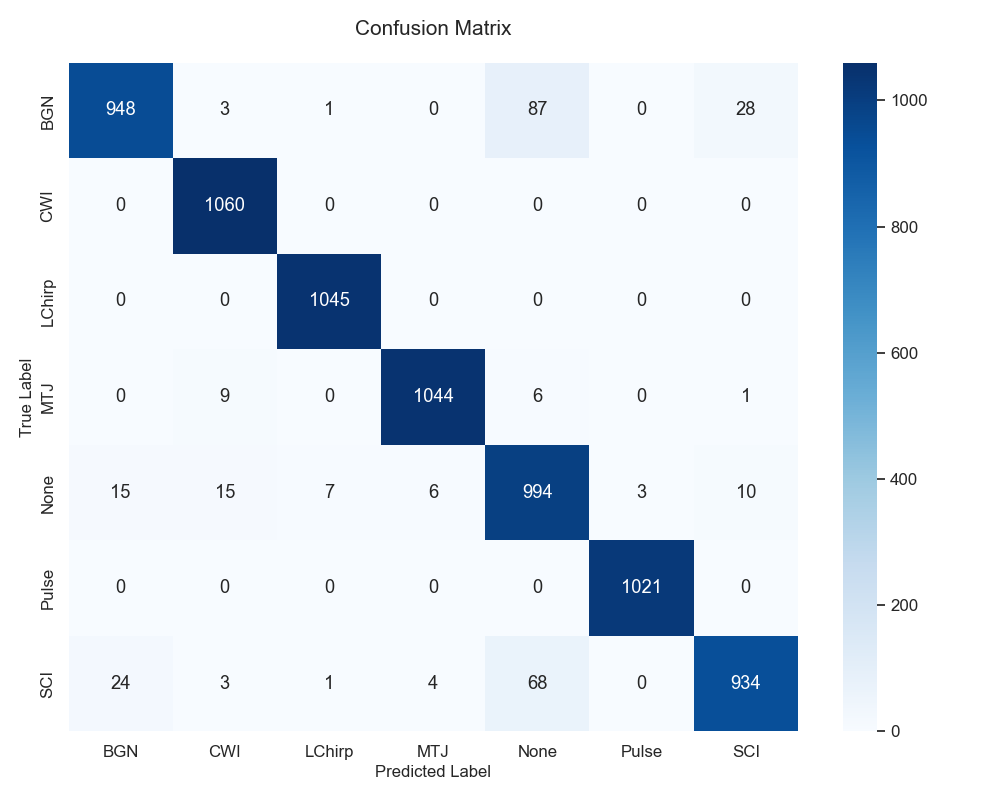}
        \caption{Confusion Matrix of the GAC-KAN model on the test dataset. The vertical axis represents the true labels, and the horizontal axis represents the predicted labels.}
        \label{fig:confusion_matrix}
    \end{figure}

To provide granular insights into the classification behavior and error distribution, Fig. \ref{fig:confusion_matrix} depicts the confusion matrix computed on the test set across all JNR levels. The evident diagonal dominance underscores the model's high precision, particularly for distinct jamming types such as CWI and MTJ, which exhibit near-zero misclassification. While the model demonstrates robust discriminability, minor confusion persists in specific scenarios: notably, 87 samples of PBNJ are misclassified as Thermal Noise ('None'), a physically consistent error attributable to their spectral similarity at low JNRs, and a small fraction of SCI is confused with PBNJ due to noise obscuring the sinusoidal signatures. Despite these localized overlaps, the proposed GAC-KAN effectively disambiguates morphologically similar signals, such as LChirp versus SCI, and maintains a substantial True Positive rate for Pulse jamming (1021 samples), thereby validating the efficacy of the multi-scale feature extraction backbone in capturing distinctive time-frequency fingerprints.

% =================================================================================
% SECTION VII: CONCLUSION
% =================================================================================
% [MODIFIED] Conclusion to reinforce the "Enabler for GenAI" narrative
\section{Conclusion}
\label{sec:conclusion}

This paper presented GAC-KAN, a physics-guided, ultra-lightweight framework designed to secure GNSS services in the era of Generative AI. By leveraging a generative simulation strategy for data synthesis, we addressed the challenge of training data scarcity. Through the novel integration of Multi-Scale Ghost-ACB blocks and a Kolmogorov-Arnold Network (KAN) head, the proposed model achieves state-of-the-art accuracy (98.0\%) while maintaining an extremely compact footprint (0.13 M parameters).

The key significance of GAC-KAN lies in its efficiency. With a computational cost of only 0.19 G FLOPs, it addresses the critical resource bottleneck in modern consumer electronics, capable of operating as an "always-on" background process. This ensures robust PNT security without compromising the performance of power-hungry Generative AI applications running on the same edge hardware. These findings position GAC-KAN as a vital enabler for the secure and reliable deployment of next-generation AI-native consumer devices.

% =================================================================================
% APPENDIX
% =================================================================================
\appendices

\section{Performance Evaluation Metrics}
\label{app:metrics}

To comprehensively assess the effectiveness of the proposed framework, we employ a diverse set of evaluation metrics. These metrics quantify the model's ability to correctly classify various jamming signals and evaluate its computational efficiency.

    \subsection{Accuracy}
    Accuracy measures the overall correctness of the model across all classes. It is defined as the ratio of correctly predicted samples to the total number of samples as follows:
    \begin{equation}
        \text{Accuracy} = \frac{TP + TN}{TP + TN + FP + FN},
    \end{equation}
    where $TP$, $TN$, $FP$, and $FN$ represent True Positives, True Negatives, False Positives, and False Negatives, respectively. While useful as a general indicator, accuracy alone may be misleading in imbalanced datasets; therefore, we complement it with class-specific metrics.

    \subsection{Confusion Matrix}
    The Confusion Matrix is a specific table layout that allows visualization of the performance of the algorithm. Each row of the matrix represents the instances in a predicted class, while each column represents the instances in an actual class (or vice versa).
    
    Let $C$ be the confusion matrix of size $K \times K$ for $K$ classes, where the element $C_{ij}$ denotes the number of samples belonging to class $j$ that were classified as class $i$. This matrix helps identify specific inter-class confusion, such as distinguishing between spectrally similar jamming types (e.g., distinguishing dense MTJ from PBNJ).

\subsection{Computational Complexity}
To assess the feasibility of deploying the proposed framework on resource-constrained edge devices, we evaluate the computational complexity using Floating Point Operations (FLOPs) and the number of learnable parameters. FLOPs quantify the theoretical computational cost of a single forward pass, serving as a critical indicator of inference latency and energy efficiency, while the parameter count determines the memory footprint required for storage. Minimizing both metrics is essential to ensure real-time, always-on interference monitoring in next-generation GNSS receivers.

%\bibliographystyle{IEEEtran}
%\bibliography{reference}

\end{document}